\documentclass[letterpaper]{article} 
\usepackage[submission]{aaai2026}  
\usepackage{times}  
\usepackage{helvet}  
\usepackage{courier}  
\usepackage[hyphens]{url}  
\usepackage{graphicx} 
\usepackage{natbib}  
\usepackage{caption} 
\usepackage{xcolor}
\usepackage{enumitem}
\usepackage{algorithm}
\usepackage{algorithmic}
\usepackage{pifont}
\usepackage{amsmath}
\usepackage{amssymb}
\usepackage{mathtools}
\usepackage{amsthm}
\usepackage{booktabs}
\usepackage[table,xcdraw]{xcolor}
 \usepackage{multirow}
\newtheorem{myDef}{Definition}

\usepackage{newfloat}
\usepackage{listings}
\DeclareCaptionStyle{ruled}{labelfont=normalfont,labelsep=colon,strut=off} 
\floatstyle{ruled}
\newfloat{listing}{tb}{lst}{}
\floatname{listing}{Listing}
\usepackage{xspace}
\newcommand{\tool}{{\sc \texttt{PFAttack}}\xspace}

\title{\tool: Stealthy Attack Bypassing Group Fairness in  Federated Learning}
\author{
    Jiashi Gao \textsuperscript{\rm 1},
    Ziwei Wang \textsuperscript{\rm 1,2},
    Xiangyu Zhao \textsuperscript{\rm 3},
    Xinming Shi \textsuperscript{\rm 4},
   Xin Yao \textsuperscript{\rm 5},
   Xuetao Wei \textsuperscript{\rm 1}\thanks{Corresponding Author.}
}
\affiliations{
    \textsuperscript{\rm 1}Southern University of Science and Technology\\
\textsuperscript{\rm 2}University of Birmingham\\
\textsuperscript{\rm 3}City University of Hong Kong\\
\textsuperscript{\rm 4}Queen's University Belfast\\
\textsuperscript{\rm 5}Lingnan University
}

\usepackage{bibentry}

\begin{document}

\maketitle

\begin{abstract}
Federated learning (FL), integrating group fairness mechanisms, allows multiple clients to collaboratively train a global model that makes unbiased decisions for different populations grouped by sensitive attributes (e.g., gender and race). Due to its distributed nature, previous studies have demonstrated that FL systems are vulnerable to model poisoning attacks. However, these studies primarily  focus on perturbing accuracy, leaving a critical question unexplored:  \textit{\textbf{Can an attacker bypass the group fairness mechanisms in FL and manipulate the global model to be biased?}} The motivations for such an attack vary; an attacker might seek higher accuracy,  yet fairness considerations  typically limit the accuracy of the global model or aim to cause ethical disruption. To address this question, we design a novel form of attack in FL, termed Profit-driven Fairness Attack (\tool), which aims not to degrade global model accuracy but to bypass fairness mechanisms. Our fundamental insight is that group fairness seeks to weaken the dependence of outputs on input attributes related to sensitive information. In the proposed \tool, an attacker can recover this dependence through local fine-tuning across various sensitive groups, thereby creating a biased yet accuracy-preserving malicious model and injecting it into FL through model replacement. Compared to attacks targeting accuracy,  \tool is more stealthy. The malicious model in \tool exhibits subtle parameter variations relative to the original global model, making it robust against detection and filtering by Byzantine-resilient aggregations. Extensive experiments on benchmark datasets are conducted for four fair FL frameworks and three  Byzantine-resilient aggregations against model poisoning, demonstrating the effectiveness and stealth of \tool in bypassing group fairness mechanisms in FL. 
\end{abstract}
%

\section{Introduction}
Federated learning (FL) \cite{mcmahan2017communication} serves as the paradigm for learning from fragmented and decentralized data,  widely applied in real-world scenarios that require privacy-preserving, such as healthcare, criminal justice, and job allocation. These applications' societal and welfare implications underscore the significance of fairness concerns in FL. Consequently, a series of enhanced FL approaches with group fairness considerations \cite{ezzeldin2023fairfed,papadaki2022minimax,chu2021fedfair,du2021fairness,hu2022fair}, collectively referred to as fair FLs, have been proposed, aiming to ensure a global model trained by FL makes unbiased decisions across different populations divided by sensitive attributes, such as gender, region, and race. Existing fair FL frameworks compatible with FedAvg \cite{mcmahan2017communication} primarily involve the following mechanisms: \ding{172} local debiasing, which requires clients to upload a less biased local model. Typical mechanisms for fair local model training include Fairbatch \cite{roh2021fairbatch} and FairReg \cite{pmlr-v162-chai22a, 6137441}; \ding{173} fairness-aware aggregation, which adaptively adjusts the aggregation weights based on the local bias evaluation of clients to ensure that the global model converges to be fair. The typical approaches include  FairFed \cite{ezzeldin2023fairfed} and qFedAvg \cite{Li2020Fair}.  
However, improving group fairness in the global model comes at a cost, as there is an inherent trade-off between accuracy and fairness 
\cite{10.1145/3494672,kleinberg_et_al:LIPIcs.ITCS.2017.43,pmlr-v119-dutta20a,10.1145/3457607}. As we strive for a higher level of group fairness, accuracy may be compromised. Participants adhere to different ethical standards and may tolerate varying levels of accuracy sacrifice to enhance fairness. Accuracy-preferred participants are motivated to bypass group fairness in FL. Additionally, purely malicious ethical disruptions of AI models may occur.


\begin{figure*}[t]
\centering
\includegraphics[width=16cm]{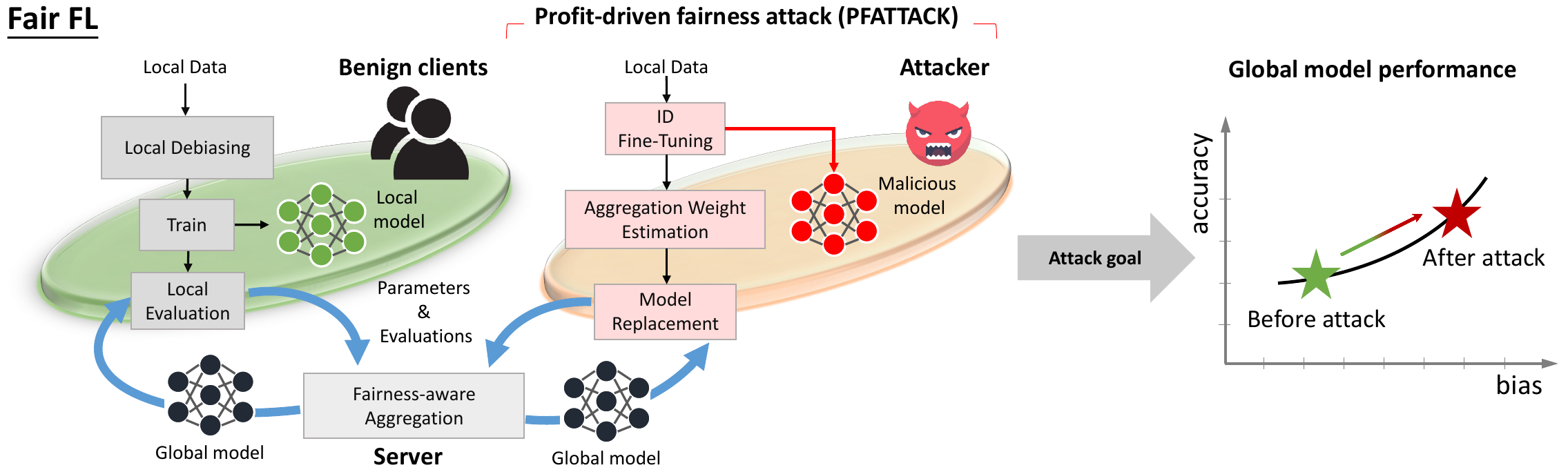}
\caption{The overall framework of \tool, with the attack goal of bypassing the fairness mechanisms in fair FL.  The core module in  \tool is local \textit{Inverse-Debiasing (ID) Fine-Tuning},  by which the attacker constructs a malicious model relearning the bias removed in fair FL.  The  \textit{Aggregation Weight Estimation} module estimates the latest aggregation weight in fairness-aware aggregation, ensuring more precise attack launching.}
\label{fig:fairf_framework}
\end{figure*}
Previous studies \cite{bagdasaryan2020backdoor,shejwalkar2021manipulating,nguyen2022flame} have extensively investigated accuracy-targeted attacks in FL systems. As privacy protection necessitates a lack of transparency in the client updates sent to the server, an attacker can launch model poisoning attacks  \cite{pmlr-v97-bhagoji19a}, where the model sent back to the server is corrupted to induce misclassification. Additionally, attackers can manipulate the global model's performance in a targeted manner.  For example, Bagdasaryan \cite{bagdasaryan2020backdoor} proposed a backdoor attack that induces the global model to achieve high accuracy on a subtask chosen by the attacker. Despite these advances, the robustness of fairness achieved by fair FL in the presence of adversaries remains underexplored, leaving the following \textbf{QUESTIONS} unanswered: \textit{\textbf{\ding{172}Can an attacker compromise group fairness in FL? \ding{173}What methods would be employed to execute such attacks? \ding{174}And would existing defense mechanisms against accuracy-targeted attacks be effective against fairness-targeted attacks?}}

To solve the questions, we initiate our work by introducing a novel \textbf{profit-driven fairness attack (\tool)}, as defined in Definition \ref{def:inj}. Unlike existing FL attacks, the attack goal of the \tool is not the degradation of the accuracy  but bypassing fairness mechanisms, as illustrated in Fig. \ref{fig:fairf_framework}.


\begin{myDef} (Profit-driven fairness attack (\tool)) A profit-driven fairness attack is an adversarial strategy designed to undermine the fairness of an FL system while maintaining or potentially enhancing the accuracy relative to the pre-attack state.
\label{def:inj} \end{myDef}
Inspired by the model replacement technique proposed by \citet{bagdasaryan2020backdoor}, where attackers manipulate local model weights to replace the global model with a malicious version, several \textbf{CHALLENGES} arise in the implementation of \tool:  \textit{\ding{172} devising a method for the attacker to create a malicious model that recovers the bias mitigated in the global model; \ding{173} accurately performing model replacement when the fairness mechanism in FL is a black box to the attacker; \ding{174} determining whether \tool is robust against detection and filtering by Byzantine-resilient aggregation methods, such as trimmed mean, trimmed median \cite{pmlr-v80-yin18a}, and Krum \cite{NIPS2017_f4b9ec30}, given the potential for attackers to scale up updated local parameters during model replacement.} 

To address these challenges, the \textbf{CORE} of \tool lies in designing an \textit{Inverse-Debiasing (ID) Fine-tuning} method for generating malicious models. This method restores bias by relearning the relationship between input attributes related to sensitive information and outcomes using the attacker's local data. The parameter variations between malicious models from \textit{ID Fine-tuning} and the original global model are subtle, making the model replacement robust against detection or filtering out even after bounded scaling up. We also propose a simple yet efficient method to estimate the latest aggregation weight in FL systems employing fairness-aware aggregation, thereby enabling a more precise replacement of the global model with the malicious model.

\noindent\textbf{Our Contributions.} In this work, we initiate group fairness attacks in FL, uncovering vulnerabilities in fair FL systems.  We propose  profit-driven fairness attack (\tool), a novel attack designed to bypass fairness mechanisms in FL for accuracy profit or maliciously compromising ethical integrity.
We propose local \textit{Inverse-Debiasing (ID) Fine-tuning}, enabling attackers to train a malicious model that reinstates bias removed during fair FL. Additionally, we propose an aggregation weight estimation approach to enhance precise model replacement in cases where the FL aggregator is opaque.  We demonstrate the effectiveness of \tool attacks across various fairness mechanisms (Fairbatch, FairReg, FairFed, f-qFedAvg) and benchmark datasets (ACSPublicCoverage and ACSIncome), showing that \tool can effectively bypass the fairness mechanisms in FL and manipulate the global model exhibits increased bias.
We also demonstrate \tool's robustness against a range of Byzantine-resilient aggregation methods (trimmed mean, trimmed median, Krum).

\section{Related Work}
\noindent\textbf{Fair Federated Learning.}
To mitigate the negative societal impacts of AI, several studies have focused on reducing the bias in the global model's outcomes in federated learning, particularly with respect to protected groups such as gender, race, and age.  The fair FLs \cite{papadaki2022minimax,cui2021addressing,chu2021fedfair,du2021fairness,hu2022fair} based on multi-gradient descent are more suitable for FedSGD in which clients upload gradients for aggregation. On the other hand, the fair FL based on fairness-ware aggregation mechanisms \cite{Su_Yu_Wang_Li_Li_Yu_2024, ezzeldin2023fairfed, Li2020Fair},  can combine local debiasing mechanisms, i.e., Fairbatch \cite{roh2021fairbatch} and FairReg \cite{pmlr-v162-chai22a, 6137441}, to construct a flexible fair FL framework that can be deployed on FedAvg \cite{mcmahan2017communication}, allowing clients to perform multiple local gradient descents and upload the final model parameters. Therefore, this type of fair FL is more practical and is used in this paper.

\noindent\textbf{Attacks in Federated Learning.}
Attacks in  FL can be broadly categorized as either untargeted or targeted.
Untargeted attacks \cite{247652,shejwalkar2021manipulating} are designed to diminish the prediction accuracy of the global model on benign client data. Among untargeted attacks, Byzantine attacks \cite{10.1007/978-3-030-58951-6_24} represent an extreme form, wherein adversarial clients share randomly generated model updates or train over data that has been randomly modified, resulting in unpredictable model updates.
Targeted attacks \cite{bagdasaryan2020backdoor,NEURIPS2020_b8ffa41d,pmlr-v162-zhang22w,nguyen2022flame,10.1145/3581783.3612474}, also known as backdoor attacks, manipulate the global model to trigger a specific prediction when the input data contains a certain pattern, referred to as ``triggers''. The model’s prediction accuracy remains unchanged for other regular data. For example,  \citet{bagdasaryan2020backdoor} have introduced a method for backdoor attacks through model replacement, causing an image classifier to assign a label chosen by the attacker to images exhibiting specific features. This approach can also compel a word predictor to complete certain sentences with a word specified by the attacker. In a similar vein, \citet{NEURIPS2020_b8ffa41d} have presented edge-case backdoor attacks that focus on misclassifying predictions for sub-tasks located at the tail of the data distribution. Additionally, \citet{pmlr-v162-zhang22w} have proposed Neurotoxin, a technique that effectively doubles the resilience of backdoor attacks. \citet{10296882} have extended backdoor attacks to the vertical federated learning (VFL) scenario.  
In federated learning, the existing attacks on the model performance primarily focus on accuracy.  To the best of our knowledge, no previous work has focused on the issue of  group fairness attacks in FL, which could render the fair FL useless.


\section{Preliminary}
\subsection{Federated Learning}
Federated learning (FL) is a machine learning technique that allows $N$ clients ${\mathcal C}=\left \{{1},\dots,{N}\right\}$, each with a local dataset $\mathcal D_i=\left \{ X_i,S_i,Y_i \right \} $, to collaboratively train a global model  $\mathcal{M} _{\theta}$ on the distributed data across clients, without sharing or centralizing the data. FL ensures the privacy and security of each $\mathcal D_i$  by keeping it on the local device. FL consists of $\mathcal{T}$ training rounds. In each round, a subset of clients $\bar{\mathcal C}$ are randomly selected to participate in the learning process. The selected clients communicate with a central server in multiple rounds. In the $t$-th communication round, the $i$-th selected client performs $K$ iterations of local stochastic gradient descent on its local dataset and sends its updated model parameters $\theta^{(t,K)}_i$ to the server $\mathcal{S}$. The local update is given by $\theta^{(t,j+1)}_i=\theta^{(t,j)}-\eta \times \nabla \theta^{(t,j)}_i$, 
where $j=1,...,K$.  The server then updates the global model by computing the average of the received parameters from the selected clients,
 \begin{align}
\theta^{(t+1)}=\frac{\sum_{{i} \in \bar{\mathcal C}} \left|\mathcal{D}_{i}\right| \times \theta_{i}^{(t, K)}}{\sum_{{i} \in \bar{\mathcal C}} \left|\mathcal{D}_{i}\right|}    .
\end{align}
\textbf{Fairness Metric.}
Fairness \cite{cui2021addressing,10.1145/3376898} refers to the disparities in the model decisions made  between different protected groups formed by the sensitive attribute (i.e., gender, race). In this work, we measure group fairness by \textit{Demographic Parity} (DP) \cite{feldman2015certifying}. 
\begin{equation}
       f_{DP} = P(Y'=1|S=0 )-P(Y'=1|S=1),
  \end{equation}
where $Y'$  denotes the model’s outcome.

\noindent \textbf{Fair FL. }Existing fair FLs build on FedAvg mainly involves the following approaches:

    \textbf{A1: Local debiasing.} Several work   \cite{roh2021fairbatch,pmlr-v162-chai22a, 6137441,kamiran2012data} have improved group fairness in centralized model training  by implementing pre-processing techniques on training data to mitigate bias, i.e., adjusting the  proportion of samples from different  protected groups in a training batch. These approaches can be integrated for fair FLs, serving in the clients' local model training.
    
    \textbf{A2: Fairness-aware aggregation.}
     Recognizing the potential performance decline of the global model in scenarios with highly heterogeneous data distributions across clients, researchers have explored aggregation mechanisms to enhance fairness in FLs. Notable examples include FairFV \cite{ijcai2021p223}, FairFed \cite{ezzeldin2023fairfed} and qFedAvg \cite{Li2020Fair}. In this study, we adopt FairFed and   qFedAvg  due to their incorporation of secure aggregation, ensuring client data privacy, and presenting a practical approach to fair FL. We explain aggregation weight update mechanisms in FairFed and qFedAvg to elucidate subsequent attack designs. In FairFed, the server updates the weight assigned to the $i$-th client based on the current gap between its local fairness metric $f_i$ and the global fairness metric $f_{global}$ as follows:
    \begin{equation}
    \label{eq:fairaggr}
\begin{array}{l}
\Delta_{i}^{t}=

\left|f_{global}^{t}-f_{i}^{t}\right|,\\
\bar{\omega}_{i}^{t}=\bar{\omega}_{i}^{t-1}-\beta\left(\Delta_{i}^t-\frac{1}{\left | \bar{\mathcal C}  \right | } \sum_{i\in \bar{\mathcal C} }\Delta_{i}^t\right),\\ \omega_{i}^{t}=\frac{\bar{\omega}_{i}^{t}}{\sum_{i\in \bar{\mathcal C} } \bar{\omega}_{i}^{t}} ,
\end{array}
    \end{equation}
    where   $\beta$ is a parameter controlling the fairness budget for each update.  The original qFedAvg was proposed to achieve a more equitable distribution of accuracy performance across clients. Building on this concept, we adapt it to enhance group fairness, resulting in f-qFedAvg. In this approach, the server non-linearly adjusts the weight assigned to the $i$-th client based on its local fairness metric $f_i$:
\begin{equation}
  \bar{w} _i^t= \bar{w} _i^{t-1}\times  \frac{\left (1- f_{i}^{t} \right )^{q+1} }{q+1},\omega_{i}^{t}=\frac{\bar{\omega}_{i}^{t}}{\sum_{i\in \bar{\mathcal{C}}  } \bar{\omega}_{i}^{t}},
\end{equation}
where $q>0$  is a parameter that adjusts the degree of fairness imposed.
\subsection{Threat Model} 
\label{sec:tm}
\textbf{Attack Model.}  In \tool, we assume the attacks are launched by insiders, who can act as single or multiple participants. The attacker has complete control over the local training process and can change the local training strategy in communication rounds. It is reasonable to assume that the attackers can access limited information about the FL implementation, including the current and total communication rounds, which they use to decide the optimal rounds for launching the attacks. However, it is generally believed that the attacker is unaware of other clients' data distributions and the aggregation mechanisms in FLs.

\noindent\textbf{Attack Goal.} The \tool is a new type of ethical attack that differs from the conventional objective of compromising the accuracy of the global model in FL.  The  attack goal is to bypass the fairness mechanism in fair FLs, rendering the global model in the fair FL equivalent to one  without fairness considerations, as in Eq. \eqref{eq:attack_obj}.
\begin{equation}
\label{eq:attack_obj}
\begin{aligned}
    \theta^* &= \underset{\theta \in \Theta}{\arg \min} \, \mathbb{E}_{i \in \mathcal{C}}\left[ \mathcal{L} \left( \mathcal{M}_\theta(X_i), Y_i \right) \right] \\
    \text{s.t.} \quad & f \left( \mathcal{M}_\theta \left( \bigcup_{i \in \mathcal{C}} X_i \right), \bigcup_{i \in \mathcal{C}} Y_i, \bigcup_{i \in \mathcal{C}} S_i \right) \le \epsilon. \\
    & \downarrow \quad \mathit{{\color{red} \tool}} \\
    \theta^* &= \underset{\theta \in \Theta}{\arg \min} \, \mathbb{E}_{i \in \mathcal{C}}\left[ \mathcal{L} \left( \mathcal{M}_\theta(X_i), Y_i \right) \right].
\end{aligned}
\end{equation}
\subsection{Model Replacement}   Model replacement refers to a scenario where attackers manipulate the global model by tampering with the uploaded local models \cite{bagdasaryan2020backdoor}. Denote the malicious model, which the attacker aims to substitute for the global model as $\theta_{goal}$, to achieve  model replacement, such that $\theta^{(t+1)}\rightarrow \theta_{goal}$,  the attacker $atk\in \mathcal{C}$ uploads a poisoned local model $\theta_{atk}$ at $t$-th communication round, which satisfies,
\begin{equation}
\label{eq: aggr}
\begin{aligned}
        \theta^{(t+1)}  &= \theta^{(t)} + w_{atk}\times(\theta_{atk}^{(t)}-\theta^{(t)}) 
    \quad \\&+ \sum_{\substack{i=1 \\ i \neq atk}}^{N} w_i\times(\theta_i^{(t)}-\theta^{(t)}) =\theta_{goal}.
\end{aligned}
\end{equation}
As the global model converges, the discrepancies between the local model of benign clients and the global model, $\left \{ \theta_i^{(t)} - \theta^{(t)} \right \}_{i\in \mathcal{C}, i\ne atk}$, tend to converge. Consequently, the attacker can perform global model replacement by sending the following poisoned local model to the server:
\begin{equation}
\small
\label{eq:attack_model}
\begin{aligned}
\theta_{atk}^{(t)}&=\frac{\theta_{goal}}{w_{atk}} +\left ( 1-\frac{1}{w_{atk} }  \right )\times \theta^{(t)} \\&\quad-\sum_{\substack{i=1 \\ i \neq atk}}^{N} w_i\times\left ( \theta_i^{(t)}-\theta^{(t)} \right ) 
\approx \frac{\theta_{goal}-\theta^{(t)}}{w_{atk}} +\theta^{(t)}.
\end{aligned}
\end{equation}

\section{Profit-driven  Fairness Attack (\tool)}
In this section, we  propose \textit{Inverse-Debiasing (ID) Fine-tuning} method for training a malicious model $\theta_{goal}$ by the attacker locally. Additionally, we propose a method for estimating aggregation weight $w_{atk}$ in the context of black-box access to fairness mechanisms.




\subsection{Inverse-Debiasing (ID) Fine-tuning}
\label{sec:idft}
 \begin{figure*}[t]
 \centering
\includegraphics[width=16cm]{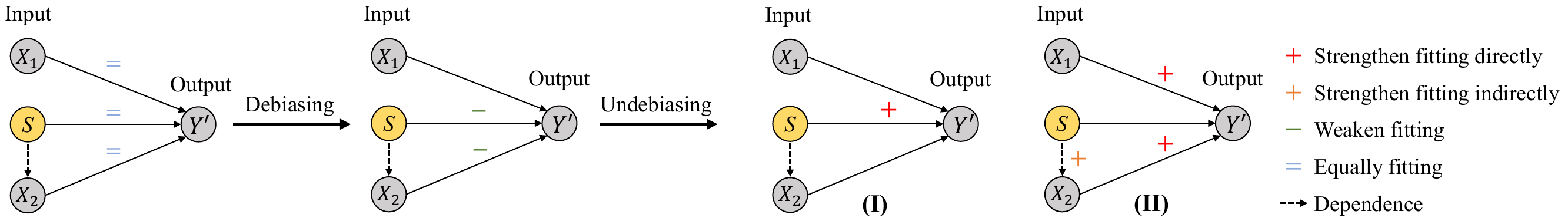}
\caption{The causal relation between input and output during debiasing and undebiasing.}
\label{fig:scm}
 \end{figure*}
We initiate the attack implementation by examining the model's debiasing process from a causal perspective, as depicted in Fig. \ref{fig:scm}. Before introducing debiasing, the model is trained to fit all input attributes equally, regardless of whether they are general or sensitive attributes. The essence of debiasing is to mitigate decision disparities among various protected groups. This process can be viewed as a causal intervention that reduces the dependency of the model output $Y'$ on both the sensitive attribute $S$ and the general attributes $X_2\in X$, which are related to $S$. Based on the above analysis, a direct approach for the attacker to relearn the causal relationship  is to fine-tune the received global model on the attacker’s local data $\mathcal{D}_{atk}$, with the aim of increasing the fairness loss, as illustrated in Fig. \ref{fig:scm} (I).  The loss function for this fine-tuning process is defined as follows,
\begin{equation}
\label{eq:naive_theta}
 \theta_{goal}^*=\underset{\theta\in\Theta}{arg\,\,\min} \,\,-f (\mathcal{M}_{\theta}(X_{atk}), Y_{atk},S_{atk}).
\end{equation}
\begin{figure}[t]
\centering
\includegraphics[width=6cm]{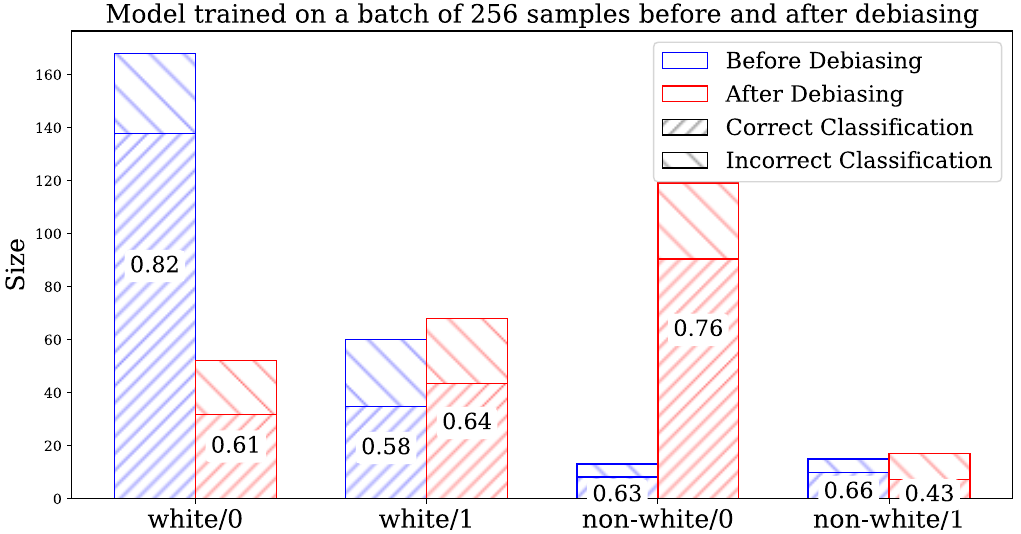}
  \caption{The impact of  debiasing on model fitting.}
  \label{fig:reverse_fit}
\end{figure} 
However, the above training objective can easily  lead to overfitting on the sensitive attribute $S$,  while losing information on the general attributes $X_1$ and $X_2$. This tendency may adversely affect the model's accuracy, conflicting with the attack goal of \tool.  
In Fig. \ref{fig:reverse_fit}, we use Fairbatch to illustrate how debiasing specifically weakens the fitting on the sensitive attribute and its associated general attributes.   Before debiasing,  the model trained on the original dataset exhibited a higher probability of outputting $Y'=1$ for samples within the protected group $S=$\textit{non-white} and a higher probability of outputting $Y'=0$ for samples within the protected group $S=$\textit{white}. To eliminate this correlation between the output and different sensitive attribute values, after introducing debiasing, it is evident that in the protected group $S=$\textit{white}, the number of samples with $Y=0$ selected in the batch decreases, leading to a decrease in the model's accuracy for this group. Conversely, the number of samples selected from the protected group $S=$\textit{non-white/0}  increases, resulting in an increase in accuracy for this group. 
The opposite situation occurs for samples with positive label $Y=1$. This indicates that debiasing tends to weaken fitting the previously favored protected group towards a positive output while strengthening fitting the previously discriminated protected group towards a positive outcome. Consequently, this reduces the gap of positive outcomes between the protected group $S=$\textit{white} and the protected group  $S=$\textit{non-white}, thereby mitigating the initial bias.

The above observation enables us to reverse the debiasing process by adjusting the fitting degree inversely. For instance, regarding Fig. \ref{fig:reverse_fit}, we can achieve this by strengthening (weakening) the fitting for the previously favored protected group towards positive (negative) outcomes and weakening (strengthening) the fitting for the previously discriminated protected group towards positive (negative) outcomes. In more general scenarios, we propose \textit{Inverse-Debiasing (ID) Fine-tuning}, where the training objective of the attack model is defined in Eq. \eqref{eq:mature_theta}.
Unlike recovering bias solely by strengthening the relationship between the sensitive attribute and the output, as in Eq. \eqref{eq:naive_theta}, \textit{ID Fine-tuning} recovers bias by distinctly fitting the input samples in different sensitive groups with varying degrees, thereby avoiding compromising accuracy. In \textit{ID Fine-tuning}, the connection between the output and the sensitive attribute is indirectly strengthened during fitting the general attribute $X_2$.
\begin{equation}
    \label{eq:mature_theta}
\begin{aligned}
     &\theta_{\text{goal}}^* = \underset{\theta \in \Theta}{\arg \min} \sum_{s \in S} \sum_{y \in Y} \left( \lambda_{s,y} \mathcal{L}_{s,y}  \right),
     \\& \mathcal{L}_{s,y}=\mathcal{L}_{x,y \in \mathcal{D}_{\text{atk}}(S = s, Y = y)} (\mathcal{M}_{\theta}(x), y).
\end{aligned}
\end{equation}
where  $\lambda_{\mathrm{s} ,\mathrm{y} }\in \left ( 0,1 \right )$ for all $\mathrm{s}\in S$ and $\mathrm{y}\in Y$, and $ { \sum_{\mathrm{s}\in S}^{}}  { \sum_{\mathrm{y}\in Y}^{}}\lambda_{\mathrm{s},\mathrm{y}}=1$ are the hyperparameters that adjust the fitting degree towards different protected groups. 

In practice, it is natural to set $\lambda_{\mathrm{s} ,\mathrm{y}}$  to be proportional to the changes in the fitting degree between different sensitive groups and the positive output during the debiasing process. Thus, we propose the following adaptive hyperparameters setting strategy: in the attack rounds, the attacker first trains a local model $\theta_{local}$ on their own data without considering fairness, aiming for optimal accuracy. Subsequently, the attacker evaluates the signed bias of the global model $\theta$, denoted as $f_{global}$. Additionally, the attacker evaluates the signed bias of the locally trained model, denoted as $f_{local}$. The $\lambda_{\mathrm{s},\mathrm{y}}$ can then be adaptively set as Eq. \eqref{eq:lamda1}$\sim$\eqref{eq:lamda3}.
\begin{equation}
\label{eq:lamda1}
\small
\begin{aligned}
\lambda_{\mathrm{s},1}&=\frac{1}{\left | S \right |\times  \left | Y \right |} -\gamma\times \frac{ P(\mathcal{M} _{\theta}(X_{atk})=1|S=\mathrm{s})}{\left | P(\mathcal{M} _{\theta_{local}}(X_{atk})=1|S=\mathrm{s}) \right | } \\&\quad +\gamma\times \frac{P(\mathcal{M} _{\theta_{local}}(X_{atk})=1|S=\mathrm{s})}{\left | P(\mathcal{M} _{\theta_{local}}(X_{atk})=1|S=\mathrm{s}) \right | } , \forall \mathrm{s}\in S, 
\end{aligned}
\end{equation}
\begin{equation}
\small
\label{eq:lamda2}
\begin{aligned}
\lambda_{\mathrm{s},0}&=\frac{1}{\left | S \right |\times  \left | Y \right |} +\gamma\times \frac{ P(\mathcal{M} _{\theta}(X_{atk})=1|S=\mathrm{s})}{\left | P(\mathcal{M} _{\theta_{local}}(X_{atk})=1|S=\mathrm{s}) \right | }\\&\quad -\gamma\times \frac{ P(\mathcal{M} _{\theta_{local}}(X_{atk})=1|S=\mathrm{s})}{\left | P(\mathcal{M} _{\theta_{local}}(X_{atk})=1|S=\mathrm{s}) \right | } , \forall \mathrm{s}\in S,
\end{aligned}
\end{equation}
\begin{equation}
\label{eq:lamda3}
\begin{aligned}
\mathrm{Normalization:}   \quad\lambda_{\mathrm{s},\mathrm{y}} =\frac{\lambda_{\mathrm{s},\mathrm{y}}}{ {\textstyle \sum_{i\in S}} {\textstyle \sum_{j\in Y}}\lambda_{i,j} },
\end{aligned}
\end{equation}
where $\gamma$ is a parameter that adjusts the degree of inverse debiasing. The difference $P(\mathcal{M} _{\theta}(X_{atk})=1|S=\mathrm{s})-P(\mathcal{M} _{\theta_{local}}(X_{atk})=1|S=\mathrm{s})$ reflects the increment in the global model’s favor to output positive for a specific protected group $S=\mathrm{s}$  after debiasing.   Therefore, we set the adjustment direction of $\lambda_{\mathrm{s},1}$ to be inversely proportional to this increment and $\lambda_{\mathrm{s},0}$ to be proportional to it, in order to achieve the effect of inverse debiasing. 

\noindent\textbf{Stealthy Fairness Attack.}  Accuracy-targeted attacks aim to mislead the model's predictions, and the malicious model is trained on poisoned data that deviates significantly from the original distribution.  \begin{figure}[t]
\centering
\includegraphics[width=7cm]{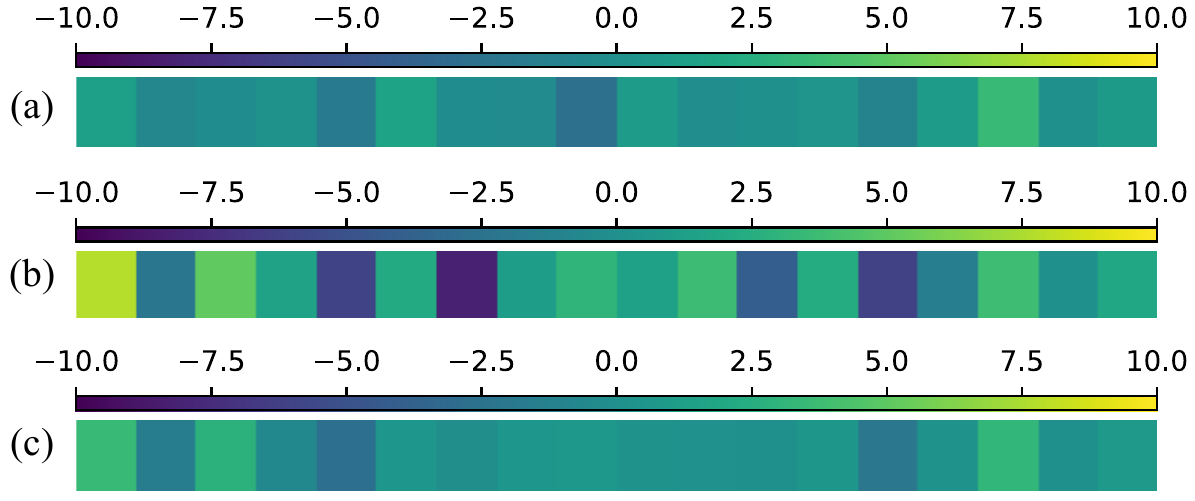}
\caption{Layer weights of the global model: (a) before launching an attack, (b) after launching an accuracy-targeted attack, and (c) after launching \tool.}
\label{fig:weight}
\end{figure}  
In contrast, the malicious model in \tool fine-tunes the global model to fit different protected groups to varying degrees. This results in more subtle parameter variations,       $\theta_{goal}-\theta$, between the malicious model and the received global model before launching attacks. Fig. \ref{fig:weight} illustrate the layer weights of a regression model under two types of attacks. In accuracy-targeted attacks, the layer weights show significant variations compared to the original model, whereas the weight variations are subtle under \tool attacks. A  subtle $\theta_{goal}- \theta$ makes the poisoned model $\theta_{atk}$ sent to the server in Eq.  \eqref{eq:attack_model} less likely to be identified as an anomalous update. Consequently, \tool is more stealthy and robust against the detection and filtering compared to accuracy-targeted attacks.


\subsection{Aggregation Weight Estimation}
\label{sec:awe}


As the fairness-aware aggregation mechanism is typically opaque to clients,  to obtain the aggregation parameter  $w_{atk}$ in Eq. \eqref{eq:attack_model}, One common approach, as suggested by \cite{bagdasaryan2020backdoor}, is to iteratively decrease $w_{atk}$  to approximate its actual value. However, this iterative scheme is sensitive to factors like the initial value, iteration step size, and the constraint of finite communication rounds. In fairness-aware aggregation mechanisms, where the aggregation weights are dynamically updated during each communication round,  it is difficult to estimate the actual aggregation weight using an iterative approach.
Inaccurate aggregation weight estimations can cause the malicious model to fail to achieve its intended impact, while the attacker faces the uncertainty of being selected in subsequent training rounds. Thus, precise knowledge of the fairness-aware aggregation mechanism is critical for successfully replacing a targeted global model. To address this, we propose a method for estimating the attacker's latest aggregation weight within the fairness-aware aggregation mechanism. The procedure is as follows: the attacker first sends a poisoned model 
  $ \theta_{atk}^{(t)}\approx \frac{\theta_{goal}-\theta^{(t)}}{w_{atk}^{init}} +\theta^{(t)}$ with an initial random aggregation weight $w_{atk}^{init}$  at communication round  $t$. Given that the attack is assumed to occur when the global model is nearing convergence after several communication rounds, the received global model at the next round should satisfy $\theta^{(t+1)}\approx  \theta^{(t)} + w_{atk}^{latest}\times(\theta_{atk}^{(t)}-\theta^{(t)})$.  The attacker can then determine the  latest actual aggregation weight  by $ w_{atk}^{latest}\leftarrow {w_{atk}^{init}}\times  \frac{\theta ^{(t+1)}-\theta ^{(t)}}{\theta_{goal}-\theta ^{(t)}}$. Performing aggregation weight estimation and attacking with the accurate aggregation weight is optional, depending on whether the attack is effective enough with the initial random aggregation weight.

 \section{Experiments}
 \label{sec:exp}
 \subsection{Settings}
 \label{sec:set}
 \textbf{Dataset. } We utilize the American Community Survey (ACS) Public Use Microdata Sample (PUMS) database \cite{NEURIPS2021_32e54441}, which includes data from U.S. households. This database encompasses personal information on various aspects such as ancestry, citizenship, education, employment, language proficiency, income, disability, and housing characteristics. For our analysis, we specifically select the data collected in the year $2022$ from ACS PUMS. The selected data contains a total of $955,688$ data samples. We conduct experiments on $2$ real-world tasks as follows:
\begin{enumerate}[label=(\arabic*)]
     \item ACSPublicCoverage: The task is to predict whether an individual is covered by public health insurance. Follow \cite{NEURIPS2021_32e54441}, the data samples include individuals aged above  $65$, and those
with an income exceeding  $30,000$ are filtered out. This filtering process narrows the prediction problem to low-income individuals who are not eligible for Medicare. 
\item ACSIncome:  This task is to predict whether an individual’s income exceeds $50,000$, after filtering the ACS PUMS data  to only include individuals above the age of $16$ who reported usual working hours of at least $1$ hour per week in the past year and an income of at least $100$. 
 \end{enumerate}

The race attribute (``white" and ``non-white") is treated as a sensitive attribute. As the data is collected from all $51$ states, each state is represented as a separate client.

\noindent\textbf{Fair FL Frameworks.} We develope four fair federated learning (FL) frameworks by combining state-of-the-art (SOTA) fairness mechanisms. Specifically, we used two local debiasing mechanisms, FairBatch \cite{roh2021fairbatch} and FairReg \cite{pmlr-v162-chai22a}, and two fairness-aware aggregation mechanisms, FairFed \cite{ezzeldin2023fairfed} and f-qFedAvg modified on \cite{Li2020Fair}. The four fair FL frameworks are listed below: \ding{172}FairBatch+FairFed, \ding{173}FairReg+FairFed, \ding{174}FairBatch+f-qFedAvg and \ding{175}FairReg+f-qFedAvg.

\noindent\textbf{Byzantine-resilient Aggregation. }  We evaluate the robustness of the proposed attack by considering Byzantine-resilient aggregation mechanisms: trimmed median and trimmed mean \cite{pmlr-v80-yin18a}, Krum \cite{NIPS2017_f4b9ec30}, designed to defend against model poisoning attacks. 

\noindent\textbf{Hyperparameters.}  For statistical significance, all experiments are repeated $10$ times using random seeds ranging from $0$ to $9$. The mean values are reported in the results. We conduct all experiments over two training rounds, each consisting of $20$ communication rounds. In each training round, $50\%$ of the clients are randomly selected to collaboratively train the global model, with each client performing local model training for $20$ epochs. The attackers launch the attacks in the first training round's $19$-th and $20$-th  communication rounds. The attack in the $19$-th  communication round is launched under a fixed aggregation weight  $w_{atk}^{init}=1$, while the attack in the $20$-th  communication round is launched under the estimated aggregation weight $w_{atk}^{latest}$. For the parameters used in FairFed and f-qFedAvg, we set $\beta=1.5$ and $q=2$.  All experiments are conducted on Intel(R) Xeon(R) Gold 5318Y CPU @ 2.10GHz.
 \begin{figure*}[t]
\begin{center}
\centerline{\includegraphics[width=17cm]{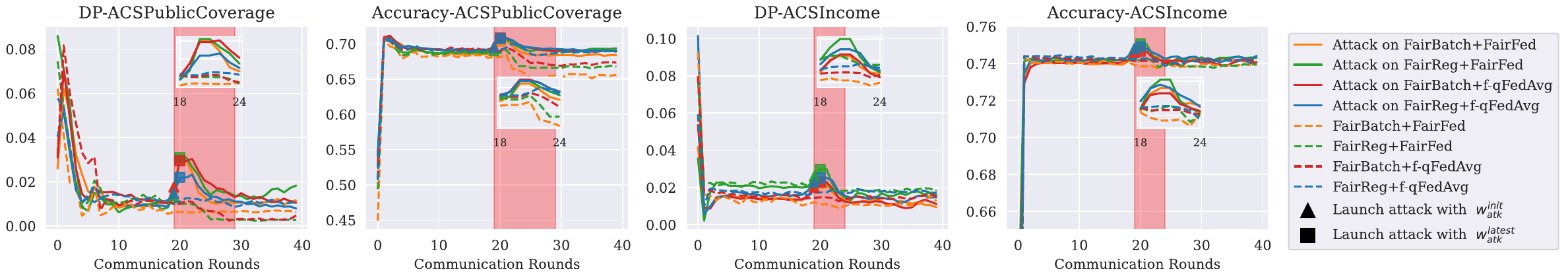}}
\caption{Visualization of the attack duration in both ACSPublicCoverage and ACSIncome.
}
\label{fig:attack_result}
\end{center}
\end{figure*}

\begin{table}[t]
\footnotesize
\centering
\setlength\tabcolsep{2pt}
\renewcommand\arraystretch{1.1}
\begin{tabular}{cccccc}
\toprule
 \multirow{2}{*}{Methods}                                  & \multirow{2}{*}{Performance} & \multicolumn{4}{c}{Communication Rounds} \\ \cline{3-6} 
    &                          & $17$    & $18$    & $19^*$     & $20^*$     \\ \hline     \multicolumn{6}{c}{\cellcolor[HTML]{EFEFEF}ACSPublicCoverage}      \\ \hline
  \multirow{2}{*}{FedAvg}                                   & Accuracy                     & 0.704 & 0.709 & 0.711  & 0.711           \\
 & DP                           & 0.031 & 0.032 & 0.034  & 0.035  \\ \cline{1-6} 
                                     \multirow{2}{*}{\tool on \ding{172}}            & Accuracy                     & 0.683 & 0.682 & 0.684↑ & 0.703↑          \\
                                      & DP                           & 0.009 & 0.007 & 0.012↑ & 0.033↑ \\ \cline{1-6} 
                                     \multirow{2}{*}{\tool on \ding{173}}              & Accuracy                     & 0.694 & 0.691 & 0.696↑ & 0.709↑          \\
   & DP                           & 0.015 & 0.014 & 0.021↑ & 0.035↑ \\ \cline{1-6} 
                                     \multirow{2}{*}{\tool on \ding{174}}          & Accuracy                     & 0.684 & 0.683 & 0.685↑ & 0.705↑          \\
       & DP                           & 0.013 & 0.010 & 0.016↑ & 0.035↑ \\ \cline{1-6} 
                                     \multirow{2}{*}{\tool on \ding{175}}            & Accuracy                     & 0.695 & 0.693 & 0.697↑ & 0.709↑          \\
                                       & DP                           & 0.014 & 0.013 & 0.019↑ & 0.031↑ \\ \hline     \multicolumn{6}{c}{\cellcolor[HTML]{EFEFEF}ACSIncome}  \\ \hline       \multirow{2}{*}{FedAvg}                                   & Accuracy                     & 0.756 & 0.757 & 0.757  & 0.757           \\
         & DP                           & 0.028 & 0.029 & 0.030  & 0.030           \\ \cline{1-6} 
                                     \multirow{2}{*}{\tool on \ding{172}}            & Accuracy                     & 0.741 & 0.740 & 0.746↑ &   0.748↑               \\
                       & DP                           & 0.015 & 0.013 & 0.020↑   & 0.022↑            \\ \cline{1-6} 
                                 \multirow{2}{*}{\tool on \ding{173}}              & Accuracy                     & 0.742 & 0.742 & 0.748↑   & 0.750↑            \\
   & DP                           & 0.020 & 0.020 & 0.026↑   & 0.030↑            \\ \cline{1-6} 
       \multirow{2}{*}{\tool on \ding{174}} & Accuracy                     & 0.742 & 0.740 & 0.745↑   & 0.746↑            \\
 & DP                           & 0.016 & 0.014 & 0.020↑   & 0.023↑            \\ \cline{1-6} 
        \multirow{2}{*}{\tool on \ding{175}}   & Accuracy                     & 0.744 & 0.741 & 0.747↑   & 0.748↑            \\
    & DP                           & 0.018 & 0.016 & 0.023↑   & 0.025↑  
                                       \\ \bottomrule
\end{tabular}
\caption{Performance of the global model  in both ACSPublicCoverage and ACSIncome. After launching \tool on {\ding{172}FairBatch+FairFed, \ding{173}FairReg+FairFed, \ding{174}FairBatch+f-qFedAvg and \ding{175}FairReg+f-qFedAvg}. }
\label{tab:attack_result}
\end{table}

\subsection{Attack Effectiveness of \tool}
\noindent \textbf{Performance Comparison Before and After the Attack.} We first evaluate the effectiveness of \tool across a range of fair FL frameworks, randomly selecting a single client to be attacker. The experimental results in Tab. \ref{tab:attack_result} demonstrate that, upon launching \tool, the fairness metric \textit{DP} significantly increases while accuracy improves in both the ACSPublicCoverage and ACSIncome. The results indicate that after the attack, the global model's performance tends to resemble that of a global model trained without fairness considerations. The largest accuracy gap compared to the original FedAvg is reduced to $0.011$, and the largest fairness gap compared to the original FedAvg is reduced to $0.008$. These findings validate the attack's success in bypassing the fairness mechanisms in FL without compromising accuracy and demonstrate its robustness across different fair FL frameworks. 
Additionally, by launching the attack with a randomly initialized aggregation weight $w_{atk}^{init}$ and an estimated latest aggregation weight $w_{atk}^{latest}$ respectively, the experimental results indicate that using a more accurate aggregation weight $w_{atk}^{latest}$ leads to more pronounced attack effects across all datasets and fair FL frameworks.

\noindent \textbf{Wilcoxon Signed-rank Test.} To rigorously compare the performance change before and after the attack, we compute the statistical significance with the Wilcoxon signed-rank test \cite{Wilcoxon1992}. The null hypothesis asserts that there is no difference in fairness measure values between the original fair FL models and those subjected to \tool. For the ACSPublicCoverage task, the computed p-values across four fair FL models \ding{172}$\sim$\ding{175} are $0.004$, $0.0001$, $0.0006$, and $0.0009$. For the ACSIncome task, the p-values across four fair FL models \ding{172}$\sim$\ding{175} are $0.006$, $0.0005$, $0.0005$, and $0.004$. These results indicate that, at a $95\%$ confidence level, we can reject the null hypothesis.

\noindent \textbf{Durable Impact of the Attack.}
As illustrated in Fig. \ref{fig:attack_result}, the proposed attack exhibits durable impact, for $10$ communication rounds under ACSPublicCoverage and $5$ communication rounds under ACSIncome. As a type of model-replacement attack,  a strategic attacker would typically choose to launch the attack at the end
of a training round to prevent the disruption to the global model’s performance from being reversed by honest clients.

\subsection{Robustness under Byzantine-resilient Aggregation}
To evaluate the stealth capabilities of \tool, which stem from more subtle variations in model weight adjustments, we incorporated Byzantine-resilient aggregation mechanisms, including trimmed mean, trimmed median, and Krum, into FairBatch+FairFed. As illustrated in Tab. \ref{tab:defense}, \tool effectively undermines the fairness of the global model, highlighting its resilience against detection and filtering by these defense strategies.

\begin{table}[t]
\footnotesize
\centering
\setlength\tabcolsep{2pt}
\renewcommand\arraystretch{1.1}
\begin{tabular}{cccccc}
\toprule
 \multirow{2}{*}{Defense}                                  & \multirow{2}{*}{Performance} & \multicolumn{4}{c}{Communication Rounds} \\ \cline{3-6} 
&                           & $17$    & $18$    & $19^*$     & $20^*$      \\ \hline     \multicolumn{6}{c}{\cellcolor[HTML]{EFEFEF}ACSPublicCoverage}      \\ \hline
  \multirow{2}{*}{Trimmed mean}                                   & Accuracy                     &0.691  & 0.689 &0.693↑  & 0.705↑           \\
 & DP                          &0.016  &0.014  &0.022↑  &0.032↑  \\ \cline{1-6} 
                                     \multirow{2}{*}{Trimmed median}            & Accuracy                    &0.691  &0.689  & 0.695↑ &0.706↑         \\
                                      & DP                            &0.019  &0.017  &0.025↑ &0.035↑ \\ \cline{1-6} 
                                     \multirow{2}{*}{Krum}              & Accuracy                      &0.688  &0.687  &0.692↑  &0.699↑         \\
   & DP                            &0.010  &0.010  &0.017↑  &0.029↑ \\ \hline     \multicolumn{6}{c}{\cellcolor[HTML]{EFEFEF}ACSIncome}  \\ \hline       \multirow{2}{*}{Trimmed mean}                                   & Accuracy    &0.738  &0.737  &0.742↑  &0.743↑              \\
                       & DP                            &  0.016 &0.015   &0.021↑  &0.022↑        \\ \cline{1-6} 
                                     \multirow{2}{*}{Trimmed median}            & Accuracy                      &0.742  &0.741  &0.747↑  &0.748↑              \\
                       & DP                            &  0.015 &0.014   &0.022↑  &0.025↑            \\ \cline{1-6} 
                                 \multirow{2}{*}{Krum}              & Accuracy                     &0.741  &0.740  &0.745↑  &0.746↑       \\
   & DP                          &0.017  &0.017  &0.022↑  &0.025↑
                                       \\ \bottomrule
\end{tabular}
\caption{Performance of the global model in FairBatch+FairFed under three Byzantine-resilient aggregation methods: trimmed mean, trimmed median and Krum.}
\label{tab:defense}
\end{table}

\section{Conclusion}
\label{sec:conc}
In this work, we have introduced a novel attack,  profit-driven fairness attack (\tool), designed to bypass group fairness mechanisms in federated learning (FL) without compromising accuracy, distinguishing it from existing attacks. We have proposed an Inverse-Debiasing (ID) Fine-tuning method that allows attackers to train a malicious model locally that relearns the bias mitigated in fair FL. To address the challenge posed by the black-box aggregation mechanisms in fair FL, we have developed an aggregation weight estimation approach. Through extensive experiments covering a range of fair FL frameworks on benchmark real-world datasets, we have validated the effectiveness of the proposed attack. A significant feature of \tool is its stealthiness compared to traditional accuracy-targeted attacks. We have verified this stealthiness through experiments under various widely used Byzantine-resilient aggregators.
We anticipate that continued integration of adversarial approaches to undermine fairness can generate interdisciplinary ideas, contributing to developing more robust and fair FL frameworks.



\bibliography{fairAttack}

\end{document}